\documentclass[letterpaper, 10 pt, conference]{ieeeconf}  % Comment this line out if you need a4paper

\IEEEoverridecommandlockouts                              % This command is only needed if 
\usepackage{xcolor}
\usepackage{graphicx}
\usepackage{subcaption}
\usepackage[margin=1in]{geometry} 
\title{\LARGE \bf

Towards Human-Robot Teaming through Augmented Reality and Gaze-Based Attention Control
}

\author{Yousra Shleibik$^{*}$, Elijah Alabi$^{*}$ and Christopher Reardon% <-this % stops a space
\thanks{$^{*}$ Equal Contribution}%
\thanks{The authors are with the Ritchie School of Engineering and Computer Science, University of Denver, Denver, CO, USA.
        {}}%
}

\begin{document}

\maketitle
\thispagestyle{empty}
\pagestyle{empty}

%%%%%%%%%%%%%%%%%%%%%%%%%%%%%%%%%%%%%%%%%%%%%%%%%%%%%%%%%%%%%%%%%%%%%%%%%%%%%%%%
\begin{abstract}

Robots are now increasingly integrated into various real world applications and domains. In these new domains, robots are mostly employed to improve, in some ways, the work done by humans.  So, the need for effective Human-Robot Teaming (HRT) capabilities grows. These capabilities usually involve the dynamic collaboration between humans and robots at different levels of involvement, leveraging the strengths of both to efficiently navigate complex situations. Crucial to this collaboration is the ability of robotic systems to adjust their level of autonomy to match the needs of the task and the human team members.

This paper introduces a system designed to control attention using HRT through the use of ground robots and augmented reality (AR) technology. Traditional methods of controlling attention, such as pointing, touch, and voice commands, sometimes fall short in precision and subtlety. Our system overcomes these limitations by employing AR headsets to display virtual visual markers. These markers act as dynamic cues to attract and shift human attention seamlessly, irrespective of the robot's physical location.

\end{abstract}
   \begin{figure}[thpb]
      \centering
      \includegraphics[scale=0.5]{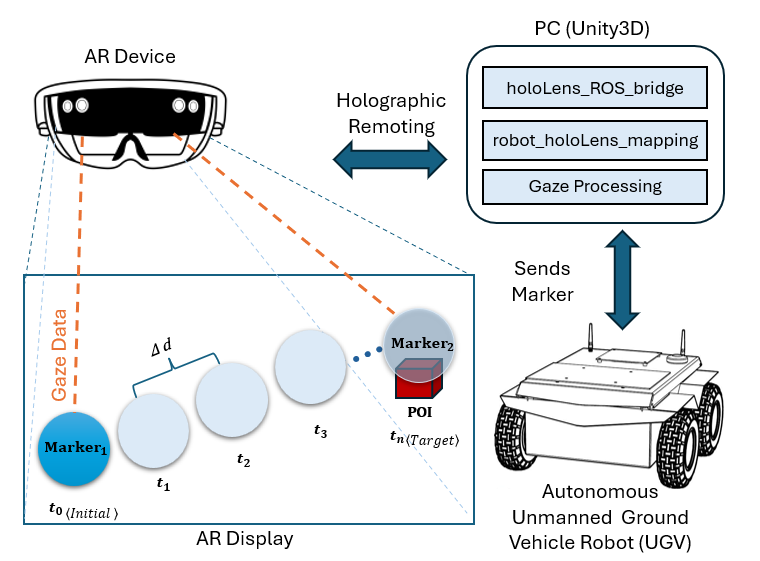}
      \caption{ The system's overall structure involves an interaction between an AR device and an autonomous unmanned ground vehicle (UGV). The UGV continuously monitors the environment and sends marker information to the AR system. The AR system waits until it detects that the user has looked at the displayed marker (Marker 1) through gaze perception from the AR device. Once this is detected, the system processes the gaze data and moves the marker to another area of interest, as shown with Marker 2. This ensures that users are effectively guided to points of interest within their environment.}
      \label{Overall}
   \end{figure}
\begin{flushleft}
\end{flushleft}

%%%%%%%%%%%%%%%%%%%%%%%%%%%%%%%%%%%%%%%%%%%%%%%%%%%%%%%%%%%%%%%%%%%%%%%%%%%%%%%%

\section{INTRODUCTION}
Attention control is the ability to seize and direct a person's focus toward a desired objective \cite{hoque2011empirical}. This concept involves purposefully shifting the person's attention toward or away from particular stimuli or tasks using persuasive strategies.
Many of such strategies have been tried and tested with mixed results. Common attention-guiding techniques such as pointing, touching and voice signal have been worked on in different contexts \cite{bodiroza2011robot,hoque2012attracting}. More recently, some of the conventional non-gaze-related methods like deictic gesture are being extended to use mixed-reality environments. A study by \cite{williams2019mixed} explores the effectiveness of the new forms of deictic gesture for multi-modal robot communication.  They  found it to be an effective communication strategy. Like the study indicated, these approaches are effective and often understood naturally. However, they lack the subtlety, flexibility and sometimes precision of gaze-based interactions.
Like deictic gestures, gaze cues are naturally used by humans. They are used to indicate interest and focus, making them a natural  and effective tool for directing attention in collaborative settings. Previous studies that used gaze-based interaction in HRI often assumed that mutual gaze \textemdash the need to look directly at each other and communicate through eye contact \textemdash is necessary \cite{moshiul2013effect, finke2005hey}. 
This assumption limits the applicability of gaze-based interaction in more dynamic and less structured environments where mutual gaze may not be possible or practical. This means that robot cannot interact with human at far distances or whenever an obstacle is blocking the line of sight needed to make eye contact. In such scenarios, this method is ineffective and needs additional mechanism to enable interaction.

We propose a HRT system (Figure \ref{Overall}) that integrates and maximizes the strength of an augmented reality (AR) device, robot and human. The HRT system places virtual visual markers visible on the AR headset at various locations depending on the specific application. These markers are dynamic cues to attract and direct the human's attention, regardless of the robot's and human physical position. By using AR technology and gaze perception, the robot can effectively guide the user's focus to specific areas (attention attraction) or divert their attention away their fixation if necessary (attention shifting), regardless of whether or not the point of interest is within the human field of view.

\section{Attention Attraction and Shared Autonomy}

In this study, we define \textbf{attention attraction} as the human reaction to a new visual stimulus, such as the sudden appearance of a marker created by the ground robot within their field of view, recognized through gaze perception. We aim to harness the natural human tendency to notice and investigate visual changes, thus capturing their attention.

Shared autonomy framework combines human input with robotic control, allowing for dynamically adjustable assistance. This approach enables a seamless collaboration, adapting the level of control based on the user's needs \cite{9501975}.

These concepts are fundamental to our stud. We use augmented reality and visual cues to explore how attention can be strategically managed in controlled settings using HRT.

Our research seeks to address the following questions:

\begin{enumerate}
    \item Can robots that are aware of human gaze effectively attract and shift attention without relying on mutual gaze, gestures, touch, or speech?
    \item How does integrating gaze-based interaction in AR environments influence user awareness and their responses to changes?
    \item To what extent can such technologies enhance the efficiency and precision of human-robot collaboration in dynamic environments?
\end{enumerate}

By exploring these questions, we aim to gain insights into the possibilities and constraints of manipulating attention through gaze-driven feedback. This knowledge will contribute to improving human-robot interaction using augmented reality, and align with the broader objectives of Variable Autonomy in Human-Robot Teaming (HRT). The goal is to develop more intuitive and responsive robotic systems that can seamlessly work with humans, adapting to their needs and the ever-changing contexts of real-world applications.

\begin{figure*}[htbp]
    \centering
    \begin{tabular}{ccc}
        \begin{minipage}[b]{0.3\textwidth}
            \centering
            \includegraphics[height=3cm]{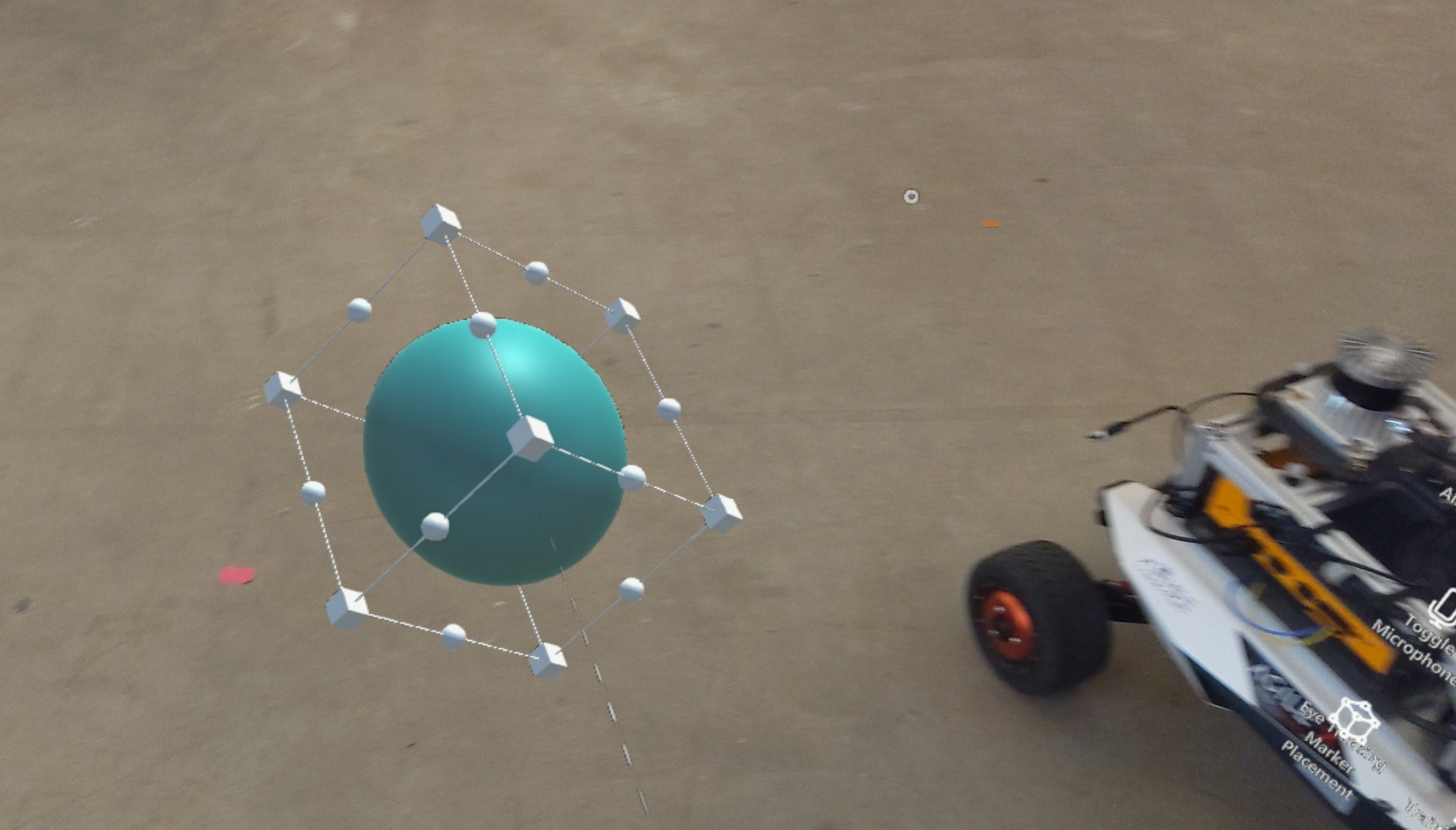}
            \\ (a)
        \end{minipage} & 
        \begin{minipage}[b]{0.3\textwidth}
            \centering
            \includegraphics[height=3cm]{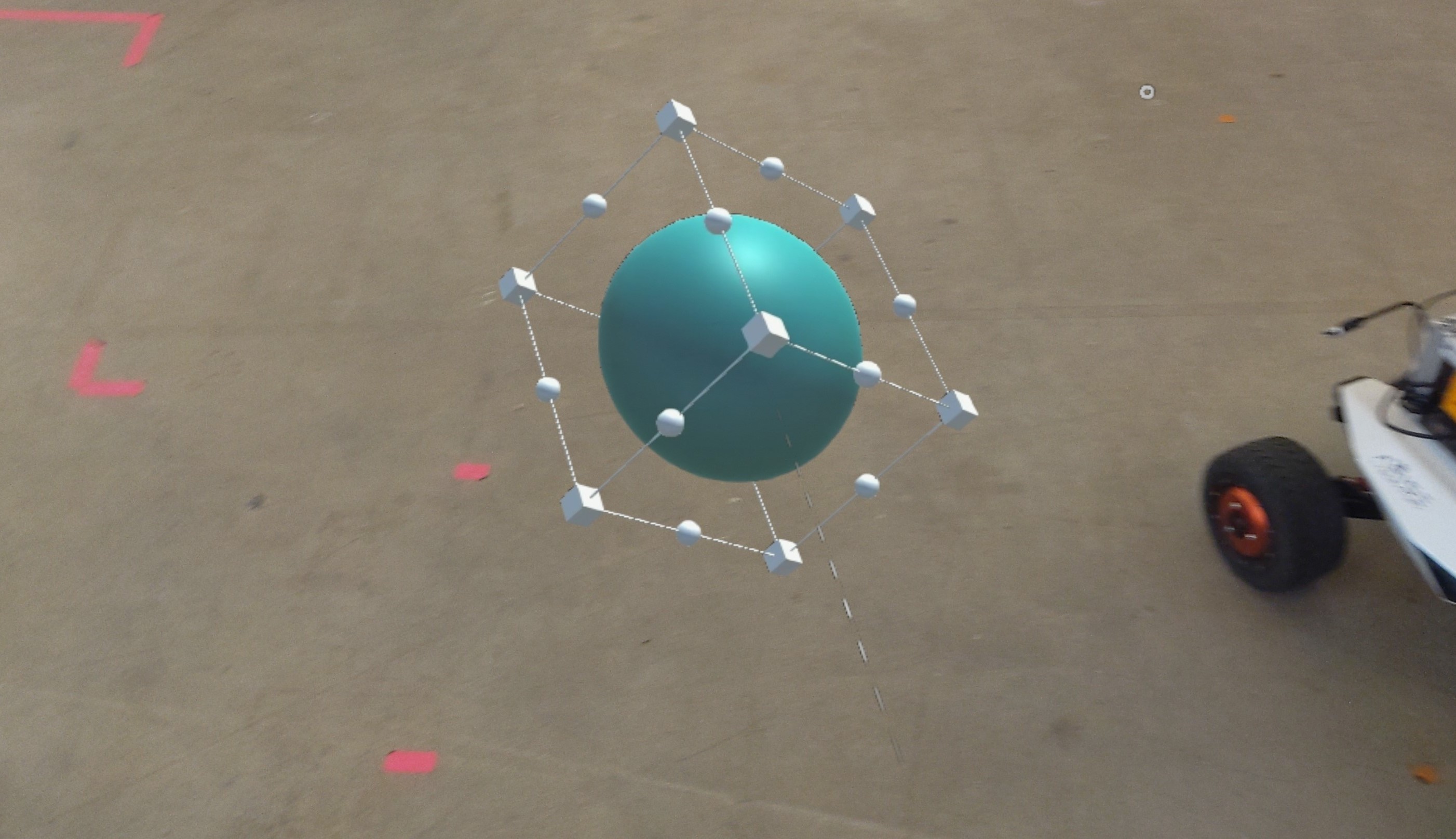}
            \\ (b)
        \end{minipage} & 
        \begin{minipage}[b]{0.3\textwidth}
            \centering
            \includegraphics[height=3cm]{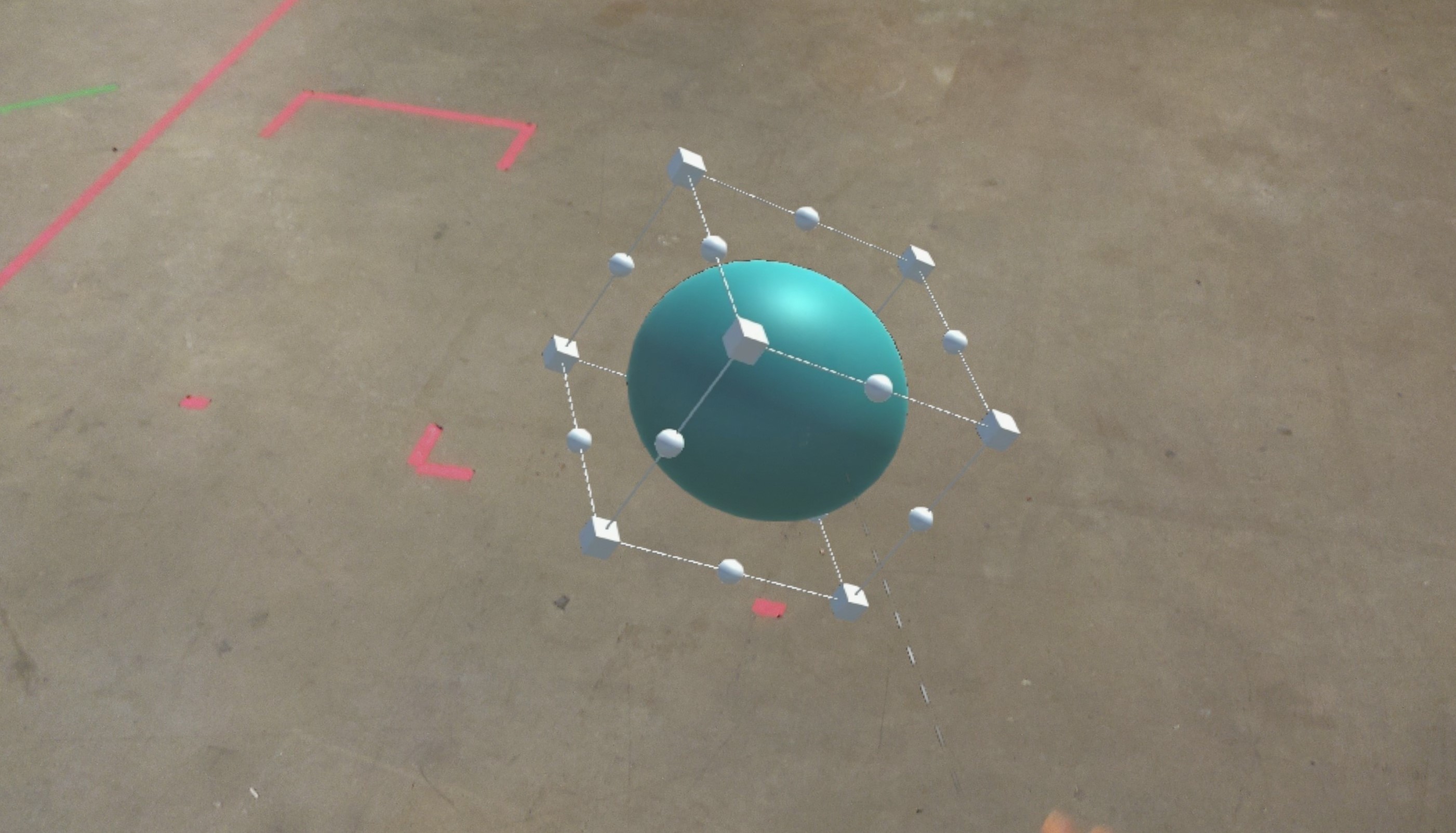}
            \\ (c)
        \end{minipage} \\
        
        \begin{minipage}[b]{0.3\textwidth}
            \centering
            \includegraphics[height=3cm]{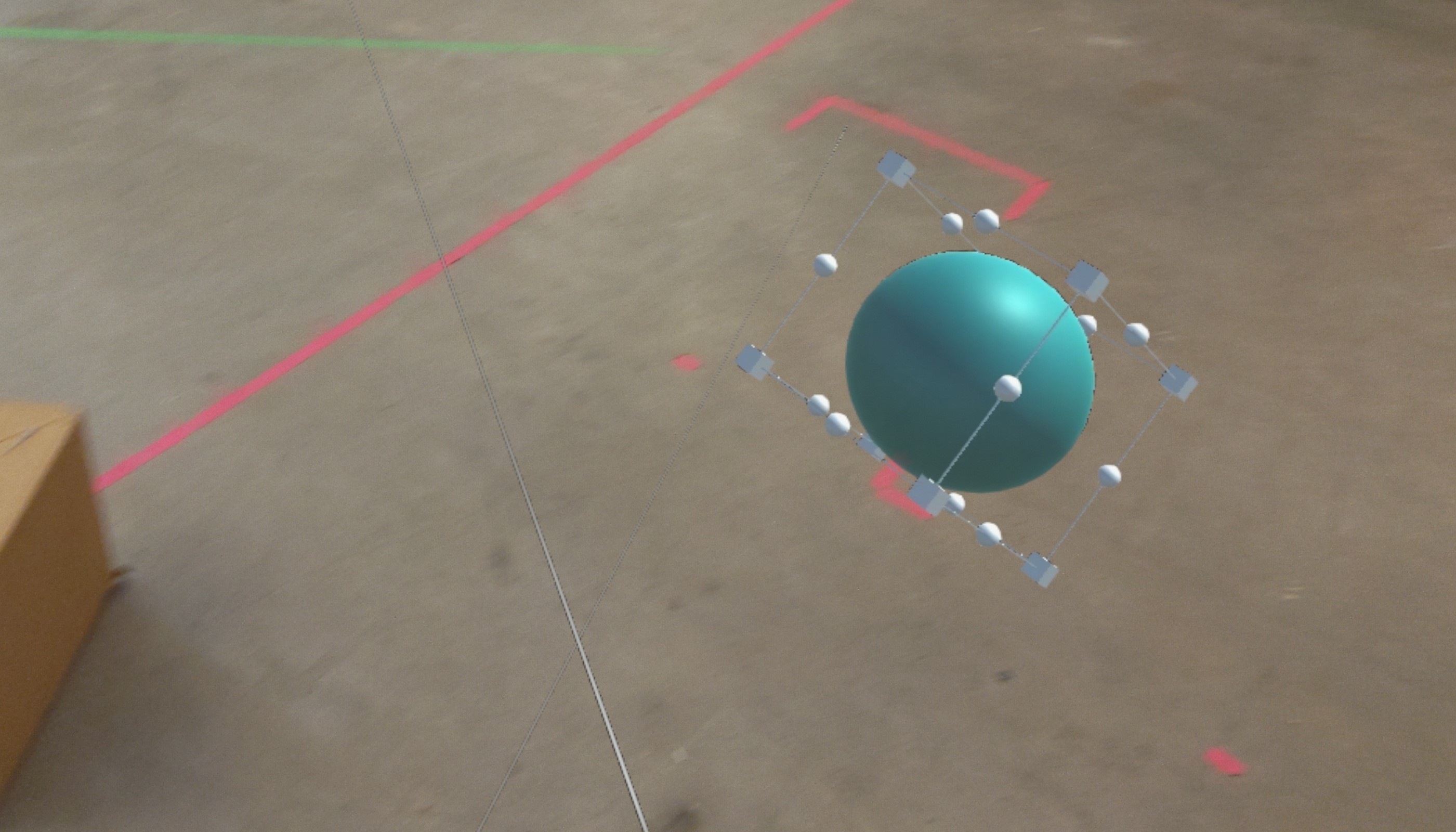}
            \\ (d)
        \end{minipage} & 
        \begin{minipage}[b]{0.3\textwidth}
            \centering
            \includegraphics[height=3cm]{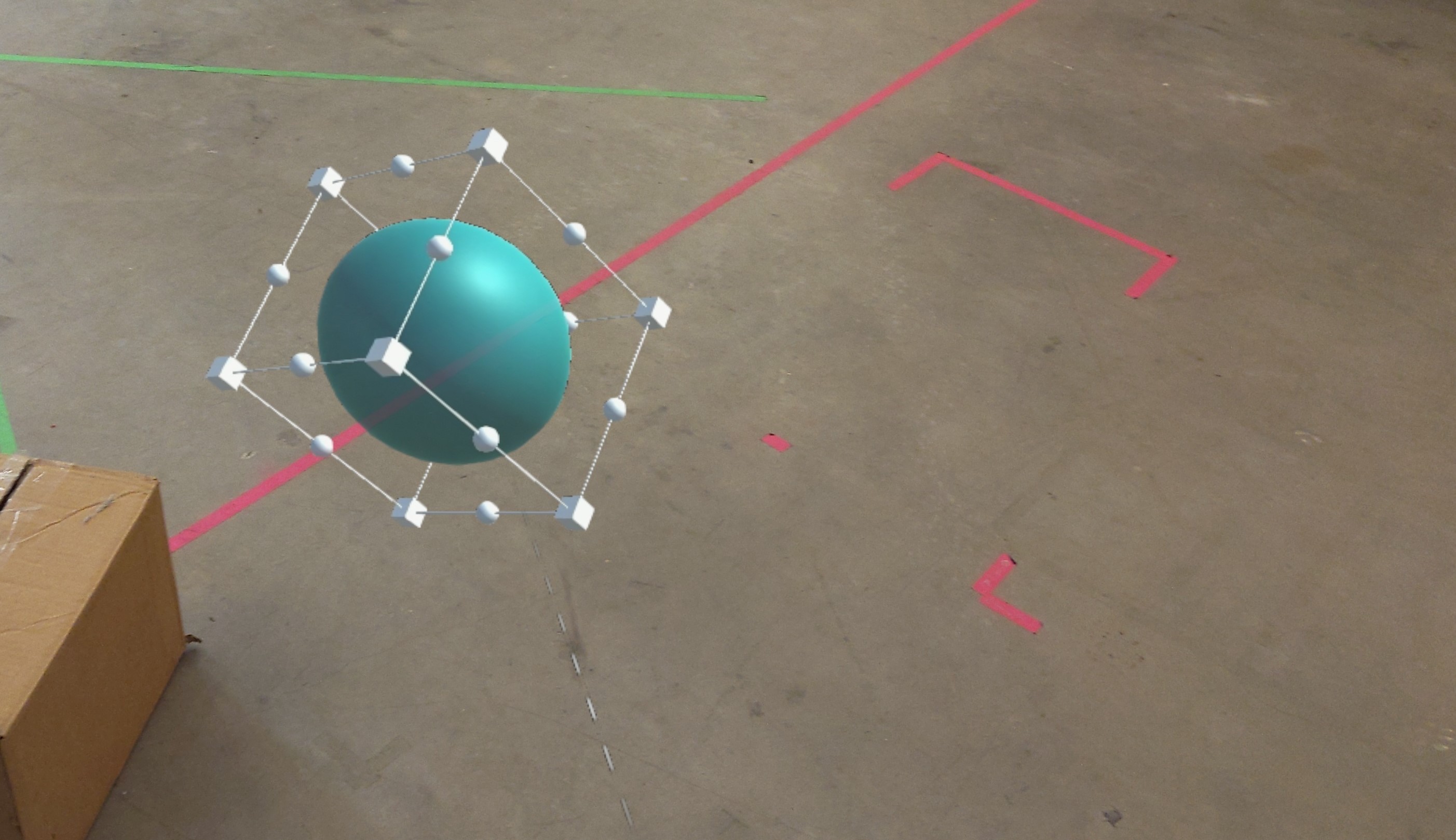}
            \\ (e)
        \end{minipage} & 
        \begin{minipage}[b]{0.3\textwidth}
            \centering
            \includegraphics[height=3cm]{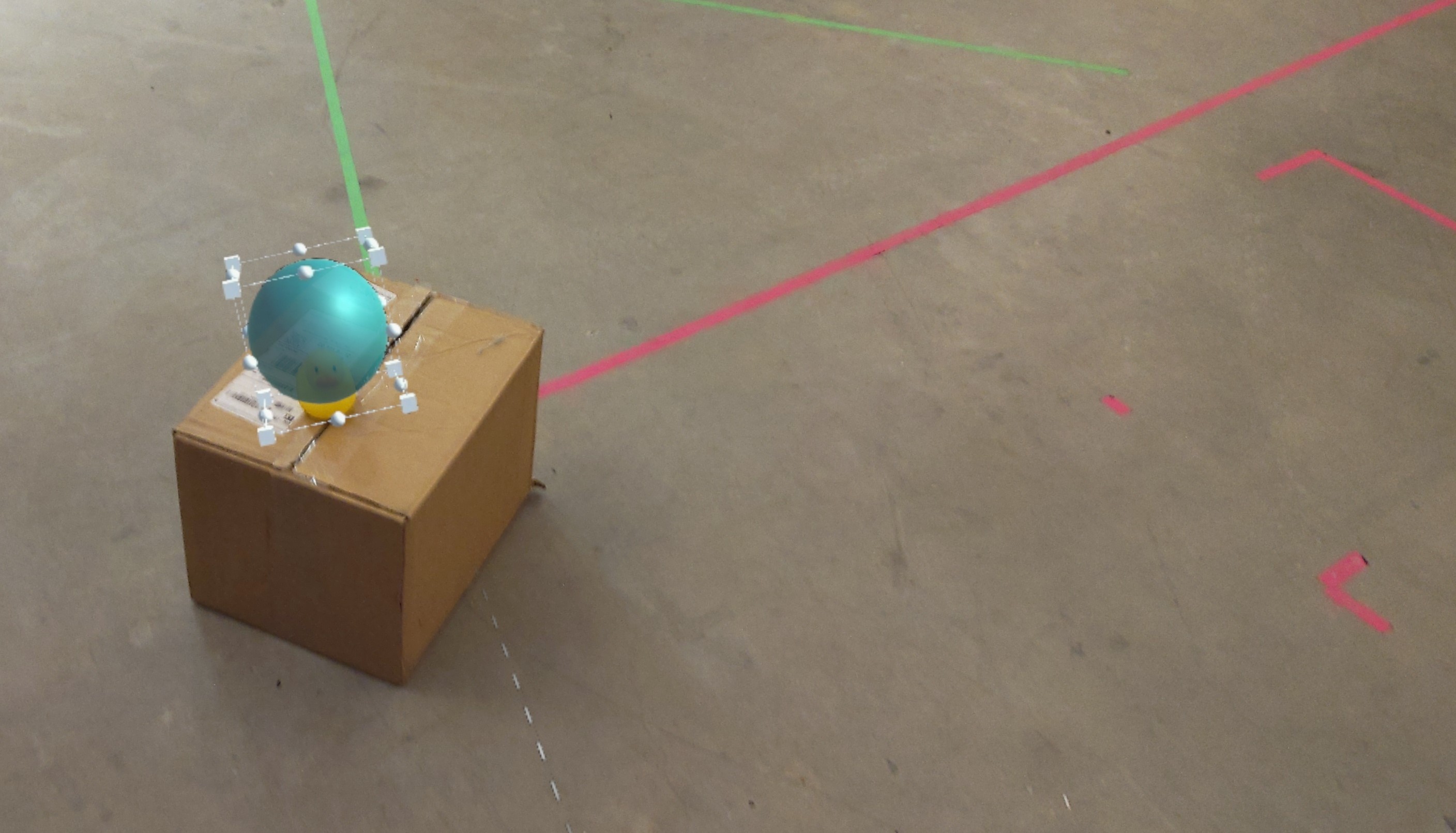}
            \\ (f)
        \end{minipage} \\
    \end{tabular}
    \caption{Attention attraction: (a) The robot sends an initial marker displayed by the AR system. (b) Once the user looks at the first marker, the system detects this and displays a second marker. (c) The system continues to show the second marker until it detects that the user is looking at it. (d) Then, it displays a third marker. (e) This process repeats, guiding the user through successive markers. (f) Finally, the user's attention is directed to the object of interest.}
    \label{fig:experiment_expectations}\vspace{-10pt}
\end{figure*}

\section{Approach}

Our approach addresses the research question by employing AR and an autonomous robot to overcome the drawbacks of traditional methods, such as physical, auditory, and basic visual cues, which often depend on proximity and direct line of sight

We propose a system where a robot uses gaze perception and an AR set to manage human attention. This follows with the principles of Variable Autonomy (VA). The system incorporates an AR device equipped with gaze tracking capabilities, projecting visual cues from a ground robot into the user's field of vision. These cues are designed either to attract the user's attention to specific stimuli or to distract them from their current focus.

The AR device monitors the user's gaze and dynamically adjusts the visual cues based on real-time analysis of attention patterns. To draw the user's gaze, the system introduces prominent AR elements, such as brightly colored shapes or moving objects, guiding the user's attention to desired locations. To distract the user, the system employs sudden, unexpected AR visuals that momentarily capture and redirect the user's focus.

In our setup, the autonomous mobile robot navigates the environment, using its sensors to identify objects of interest. Upon detecting something noteworthy, the robot sends a marker to the AR system, which displays it as a visual cue through the AR set. The system then tracks the user's gaze to confirm attention on the marker and subsequently moves the marker to another point of interest, continuously guiding the user's focus.

This approach exemplifies VA by enabling the robot to adjust its level of intervention based on the user's attention patterns, thus optimizing the collaboration process. By utilizing AR to provide real-time visual feedback based on gaze data, our system presents a solution for directing user attention in dynamic environments. Figure \ref{Overall} illustrates the system’s overall structure.

In our framework, "points of interest" (POIs) are defined flexibly to suit various application scenarios. Rather than a fixed definition, we consider it a variable that identifies POIs based on the specific context and goals of the human-robot interaction.

\section{Experimentation and Data Collection}

\subsection{System Overview}
The proposed system aims to improve upon traditional methods for directing human attention. The setup involves an AR head-mounted device (AR-HMD), specifically the HoloLens 2, paired with an autonomous robot.
\begin{itemize}
    \item \textbf{HoloLens 2 AR-HMD:} This device provides participants with a visual interface that displays markers. It continuously streams gaze tracking data from the device to the robot.

    \item \textbf{Autonomous Unmanned Ground Vehicle (UGV):} This robot strategically places visual markers within the environment. It aligns its SLAM-based map with the AR-HMD map in real-time~\cite{reardon2020enabling}, enabling it to position markers that correspond to robot-detected points of interest anywhere in the user's field of view and adjust these placements based on the user's movements and gaze direction.
\end{itemize}

\subsection{Experiment Design}

We propose to evaluate the functionality of our system through a study in which participants interact with a ground robot while wearing the AR-HMD in a controlled environment. The study will be structured into a sequence of operations designed to explore variable autonomy in human-robot teaming.

The process will begin with the \textit{Robot Interaction} phase, where the ground robot autonomously navigates the space. During this phase, the robot will place AR markers at specific points of interest, designed to either capture or shift the participant's attention as needed. These points of interest depends on the specific application of the system. This initial phase allows the robot to autonomously manage interaction and responsiveness in real-time.

Following this will be the \textit{Attention Tracking} phase. As participants engage with their environment, the robot will continuously monitor their gaze. The robot will dynamically adjust its level of autonomy based on real-time gaze data, identifying the time $t_i$ it takes for the gaze to align with each new marker $M_i$ and analyzing the patterns of gaze movement \( G(t) \) as attention shifts between different stimuli. This adaptive approach illustrates the concept of variable autonomy, as the robot adjusts its behavior based on the participant's attention feedback.

For attention tracking, we define the following variables to investigate the system's adaptability and effectiveness in a variable autonomy framework:

\textbf{Independent Variables:}
\begin{enumerate}
    \item \textbf{Distance \( \Delta d \)}: The Euclidean distance between successive dynamic markers \( M_i \) and \( M_{i+1} \). This variable is manipulated to study its effect on gaze patterns.
    %\[
    %D = \sqrt{(x_{i+1} - x_i)^2 + (y_{i+1} - y_i)^2 + (z_{i+1} - z_i)^2}
    %\]

    \item \textbf{Time Interval \( \Delta t \)}: The interval between movements of the marker. This variable is adjusted based on experimental conditions to observe its impact on gaze behavior.
    %\[
    %\Delta t = f(t_i, D, G(t))
    %\]
\end{enumerate}

\textbf{Dependent Variable:}
\begin{enumerate}
    \item \textbf{Gaze Movement \( G(t) \)}: The trajectory and speed of gaze movement.
    \[
    G(t) = \left( \frac{dx(t)}{dt}, \frac{dy(t)}{dt}, \frac{dz(t)}{dt} \right)
    \]
\end{enumerate}

The time interval for moving the markers is determined by a function \( f \) that incorporates feedback from the gaze tracking data.\( f \) depends on the application of the system. This function allows the system to dynamically adjust marker placement based on real-time analysis of the participant's gaze patterns, embodying the principles of variable autonomy. This is is important for understanding how effectively the robot can use AR markers to guide the participant's attention, thereby demonstrating the robot's ability to adapt its behavior based on human responses.

We utilize Raycasting to precisely determine if the user is gazing at the augmented reality (AR) environment's displayed marker. Raycasting is a technique for projecting an invisible ray from the user's perspective into the 3D space of an AR scene. Calculating the intersection of this ray with items in the scene allows us to establish if the user's gaze coincides with the marker. Specifically, we cast a ray from the user's eye position, as determined by HoloLens gaze data, in the direction of their gaze vector. The Raycast equation, as stated by \cite{glassner1989introduction} can be expressed as follows :

\[
\mathbf{P} = \mathbf{O} + t \cdot \mathbf{D} 
\]

Where \( \mathbf{P} \) is the point of intersection, \( \mathbf{O} \) is the origin of the ray (the user's eye position), \( \mathbf{D} \) is the direction of the gaze vector, and \( t \) is a scalar value representing the distance from the origin to the intersection point. If the ray intersects with the marker's bounding box, we confirm that the user is looking at the marker. This detection mechanism is crucial for our system, as it triggers the next step of moving the marker to a new area of interest once the user has acknowledged the current marker, thus ensuring a dynamic and interactive user experience.\\

Finally, in the \textit {Data Collection} phase, we gather and analyze the data accumulated from the gaze tracking. Metrics like gaze alignment, the time required to focus on new markers, and the movement patterns of gaze shifts are collected.

\section{Initial results}

In our initial experiments, conducted on two users,  we successfully confirmed that the system could detect when users looked at the marker. The experiment focused on attention attraction, where the robotic system collaborated with the human user to guide their attention effectively. Upon detecting that the user’s gaze was fixed on the marker, the system accurately updated the marker's position to a new area of interest within the user's field of view, demonstrating the effectiveness of the gaze-guided interaction. 

The robotic system operated under a shared autonomy framework, meaning it waited for the user's response—explicitly tracking the user's gaze on the marker or visual cue— before generating another marker to shift the user's attention to that new marker. This interaction ensured that users were promptly notified of the point of interest in their surroundings, increasing their situational awareness. Figure \ref{fig:experiment_expectations} illustrates an example of marker update interaction for attention attraction.

We plan to experiment with more users to gather comprehensive data and further validate our approach. Additionally, we aim to conduct more experiments focused on attention shifting to refine the interaction model. In future iterations, we intend to have the robot itself perform the ray cast calculations independently, which could streamline the process and significantly enhance the system's efficiency.

\section{Conclusion and Future work:}

In this paper, we introduce a new approach to attention control in human-robot interaction (HRI) using an augmented reality (AR) headset and a mobile robot. This system differs from traditional methods like audio and visual cues. Our approach leverages real-time gaze tracking to adjust AR markers based on user attention, ensuring precise and personalized interaction. We present an experimentation plan that will allow us to assess gaze tracking accuracy, participant engagement, and repeatability.

Based on related studies, we believe that our human-AR-robot system has the potential to overcome some limitations common in previous attention control studies, effectively direct human attention, improve user interaction, and advance both experimental and practical applications in HRI.

%%%%%%%%%%%%%%%%%%%%%%%%%%%%%%%%%%%%%%%%%%%%%%%%%%%%%%%%%%%%%%%%%%%%%%%%%%%%%%%%

%%%%%%%%%%%%%%%%%%%%%%%%%%%%%%%%%%%%%%%%%%%%%%%%%%%%%%%%%%%%%%%%%%%%%%%%%%%%%%%%

%%%%%%%%%%%%%%%%%%%%%%%%%%%%%%%%%%%%%%%%%%%%%%%%%%%%%%%%%%%%%%%%%%%%%%%%%%%%%%%%

%%%%%%%%%%%%%%%%%%%%%%%%%%%%%%%%%%%%%%%%%%%%%%%%%%%%%%%%%%%%%%%%%%%%%%%%%%%%%%%%
\bibliographystyle{ieeetr}
\bibliography{Bibliography}

\begin{thebibliography}{1}

\bibitem{hoque2011empirical}
M.~M. Hoque, T.~Onuki, E.~Tsuburaya, Y.~Kobayashi, Y.~Kuno, T.~Sato, and S.~Kodama, ``An empirical framework to control human attention by robot,'' in {\em Computer Vision--ACCV 2010 Workshops: ACCV 2010 International Workshops, Queenstown, New Zealand, November 8-9, 2010, Revised Selected Papers, Part I 10}, pp.~430--439, Springer, 2011.

\bibitem{bodiroza2011robot}
S.~Bodiroza, G.~Schillaci, and V.~V. Hafner, ``Robot ego-sphere: An approach for saliency detection and attention manipulation in humanoid robots for intuitive interaction,'' in {\em 2011 11th IEEE-RAS International Conference on Humanoid Robots}, pp.~689--694, IEEE, 2011.

\bibitem{hoque2012attracting}
M.~M. Hoque, T.~Onuki, D.~Das, Y.~Kobayashi, and Y.~Kuno, ``Attracting and controlling human attention through robot's behaviors suited to the situation,'' in {\em Proceedings of the seventh annual ACM/IEEE international conference on human-robot interaction}, pp.~149--150, 2012.

\bibitem{williams2019mixed}
T.~Williams, M.~Bussing, S.~Cabrol, E.~Boyle, and N.~Tran, ``Mixed reality deictic gesture for multi-modal robot communication,'' in {\em 2019 14th ACM/IEEE International Conference on Human-Robot Interaction (HRI)}, pp.~191--201, IEEE, 2019.

\bibitem{moshiul2013effect}
M.~Moshiul~Hoque, T.~Onuki, Y.~Kobayashi, and Y.~Kuno, ``Effect of robot’s gaze behaviors for attracting and controlling human attention,'' {\em Advanced Robotics}, vol.~27, no.~11, pp.~813--829, 2013.

\bibitem{finke2005hey}
M.~Finke, K.~L. Koay, K.~Dautenhahn, C.~L. Nehaniv, M.~L. Walters, and J.~Saunders, ``Hey, i'm over here-how can a robot attract people's attention?,'' in {\em ROMAN 2005. IEEE International Workshop on Robot and Human Interactive Communication, 2005.}, pp.~7--12, IEEE, 2005.

\bibitem{9501975}
M.~Selvaggio, M.~Cognetti, S.~Nikolaidis, S.~Ivaldi, and B.~Siciliano, ``Autonomy in physical human-robot interaction: A brief survey,'' {\em IEEE Robotics and Automation Letters}, vol.~6, no.~4, pp.~7989--7996, 2021.

\bibitem{reardon2020enabling}
C.~Reardon, J.~Gregory, C.~Nieto-Granda, and J.~G. Rogers, ``{Enabling Situational Awareness via Augmented Reality of Autonomous Robot-Based Environmental Change Detection},'' in {\em International Conference on Human-Computer Interaction: Virtual, Augmented, and Mixed Reality}, pp.~611--628, Springer, 2020.

\bibitem{glassner1989introduction}
A.~S. Glassner, {\em An introduction to ray tracing}.
\newblock Morgan Kaufmann, 1989.

\end{thebibliography}

\end{document}